\documentclass{article}

\usepackage{tikz}
\usepackage{microtype}
\usepackage{svg}
\usepackage{graphicx}
\usepackage[table]{xcolor}
\usepackage{tabularx}
\usepackage{adjustbox}
\usepackage{subcaption}
\usepackage{booktabs} 
\usepackage{multirow}
\usepackage{stmaryrd}
\usepackage{hyperref}
\usepackage{tabularx}
\usepackage[most]{tcolorbox}
\usepackage{fvextra} 
\usepackage{listings}
\usepackage[table]{xcolor}

\definecolor{headerblue}{RGB}{225,235,247}

\definecolor{systemdark}{RGB}{223,238,249}
\definecolor{systemlight}{RGB}{241,247,252}
\definecolor{systemTitleRed}{RGB}{248,205,205}
\definecolor{taskTitleOrange}{RGB}{252,222,185}
\definecolor{taskdark}{RGB}{226,242,232}
\definecolor{tasklight}{RGB}{243,249,245}

\definecolor{linegray}{gray}{0.75}
\newtcolorbox{promptbox}[1][]{
  breakable,
  enhanced,
  colback=gray!3,
  colframe=gray!45,
  boxrule=0.35pt,
  arc=2pt,
  outer arc=2pt,
  left=5pt,
  right=5pt,
  top=7pt,
  bottom=5pt,
  before skip=4pt,
  after skip=4pt,
  fontupper=\ttfamily\small,
  halign=left,
  fonttitle=\bfseries\small,
  coltitle=black,
  attach boxed title to top left={
    yshift=-2mm,
    xshift=5pt
  },
  boxed title style={
    colback=white,
    colframe=gray!45,
    boxrule=0.35pt,
    arc=1pt,
    outer arc=1pt,
    left=4pt,
    right=4pt,
    top=1pt,
    bottom=1pt
  },
  title={#1}
}
\usepackage{enumitem}
\usepackage{wrapfig}

\usepackage{subcaption}

\usepackage[accepted]{icml2026}

\usepackage{amsmath}
\usepackage{amssymb}
\usepackage{mathtools}
\usepackage{amsthm}

\usepackage[capitalize,noabbrev]{cleveref}

\theoremstyle{plain}

\theoremstyle{definition}

\theoremstyle{remark}

\usepackage[textsize=tiny]{todonotes}

\icmltitlerunning{Benchmarking and Error Analysis of Deep Search and Deep Research LLM Agents}

\begin{document}

\twocolumn[
  \icmltitle{Benchmarking and Error Analysis of \\ Deep Search and Deep Research LLM Agents}



  \icmlsetsymbol{equal}{*}

  \begin{icmlauthorlist}
    \icmlauthor{Zixuan Yu}{equal,sysu}
    \icmlauthor{Zhenheng Tang}{equal,hkust}
    \icmlauthor{Xiaowen Chu}{hkustgz}
    \icmlauthor{Bo Han}{hkbu}
  \end{icmlauthorlist}

  \icmlaffiliation{sysu}{School of Computer science and engineering, Sun Yat-sen University}
  \icmlaffiliation{hkust}{CSE, The Hong Kong University
of Science and Technology}
  \icmlaffiliation{hkustgz}{DSA Thrust, The Hong Kong University of Science and Technology (GuangZhou)}
  \icmlaffiliation{hkbu}{TMLR Group,
Department of Computer Science, Hong Kong Baptist University}

  \icmlcorrespondingauthor{Bo Han}{bhanml@comp.hkbu.edu.hk}

  \icmlkeywords{Machine Learning, ICML}

  \vskip 0.3in
]



 \printAffiliationsAndNotice{\icmlEqualContribution}

\begin{abstract}
As Large Language Models (LLMs) evolve into autonomous Deep Search and Deep Research Agents capable of executing long-horizon, multi-step tasks, evaluating their reliability becomes increasingly critical. However, existing evaluation paradigms predominantly focus on final task outcomes, leaving a significant gap in understanding the intermediate execution traces, where agents are highly susceptible to various errors. To address this critical gap, we introduce a comprehensive methodology and analytical framework designed specifically to diagnose errors in intermediate workflows. Building upon this framework, we establish a structured taxonomy that categorizes these vulnerabilities into two primary dimensions: \textbf{System-Level Errors} and \textbf{Task-Level Errors}. Through a systematic examination of step-by-step execution traces and agent-environment interactions, we reveal how these underlying errors affect overall performance. Finally, through empirical experiments and systematic analysis of the identified error types, we provide actionable design guidelines, aiming to enhance the robustness, reasoning capabilities, and reliability of Deep Search and Deep Research agents.
\end{abstract}

\section{Introduction}
Recent advancements in Large Language Models (LLMs) \cite{achiam2023gpt,yang2025qwen3} have catalyzed a pivotal shift from passive text generators to autonomous agents capable of sustained interaction with external environments \cite{guo2024large, luo2025large, schick2023toolformer, yao2022react}. At the forefront of this evolution are two distinct yet equally critical agentic paradigms designed to address knowledge-intensive tasks: \textbf{Deep Search} \cite{xi2025survey,li2025websailor} and \textbf{Deep Research} \cite{xu2025comprehensive} agentic frameworks. While \textbf{Deep Search} agents emphasize dynamic, long-horizon information acquisition, iteratively navigating the unstructured web to identify specific facts, \textbf{Deep Research} agents focus on synthesizing large-scale, multi-source context into comprehensive analytical reports. As these frameworks are increasingly deployed in complex real-world scenarios, understanding their structural bottlenecks and execution limitations has emerged as an important research priority \cite{zhou2023webarena,wei2025browsecomp}.

However, prevailing benchmarks \cite{mialon2023gaia, wei2025browsecomp} often obscure the true capabilities of these agents by relying primarily on end-to-end performance metrics. This outcome-oriented paradigm introduces a significant blind spot in agent evaluation, as a correct response may not arise from a valid reasoning process, while an incorrect outcome offers limited insight into the underlying architectural deficiencies or sources of failure. Unlike single-turn interactions, LLM agents are highly susceptible to complex, cascading errors throughout their iterative workflows \cite{zhu2025llm}. Although recent studies have begun to examine intermediate reasoning trajectories and process-level behaviors \cite{mohammadi2025evaluation}, systematic investigations remain limited, particularly for deep-search and deep-research agents that operate over extended, multi-stage information-seeking workflows.

To address this gap, we formalize a two-level error perspective for agent evaluation. \textbf{System-Level errors} are shared across both Deep Search and Deep Research, including malformed tool calls, schema violations, environment rejections, and infrastructure failures (e.g., timeouts and rate limits). \textbf{Task-Level errors} are paradigm-specific and reflect the cognitive objective of each workflow.

For \textbf{Deep Search}, task-level failures primarily arise from incomplete exploration, weak evidence accumulation, and premature termination. For \textbf{Deep Research}, task-level failures primarily arise from long-context synthesis weaknesses, such as poor cross-source reconciliation and insufficient citation-grounded reasoning. We intentionally defer fine-grained category definitions to the taxonomy section.

Building on this formulation, we present a unified error diagnosis framework and benchmark for process-level evaluation from full execution traces. Based on this framework, we derive a structured taxonomy spanning both levels and both paradigms, and then conduct an empirical study to characterize how error types emerge and propagate across agent workflows. Our main contributions are as follows:

\begin{itemize}
\item  We introduce a process-oriented evaluation framework that captures full agent trajectories, enabling reproducible diagnosis beyond end-to-end accuracy.
\item We propose a structured taxonomy that separates shared \textbf{System-Level} failures from paradigm-specific \textbf{Task-Level} failures in Deep Search and Deep Research.
\item We perform trace-based analysis to quantify how different error types occur and propagate, and we summarize actionable implications for improving agent robustness and reliability.
\end{itemize}

\begin{figure*}[htbp]
    \centering
    \includegraphics[width=\textwidth]{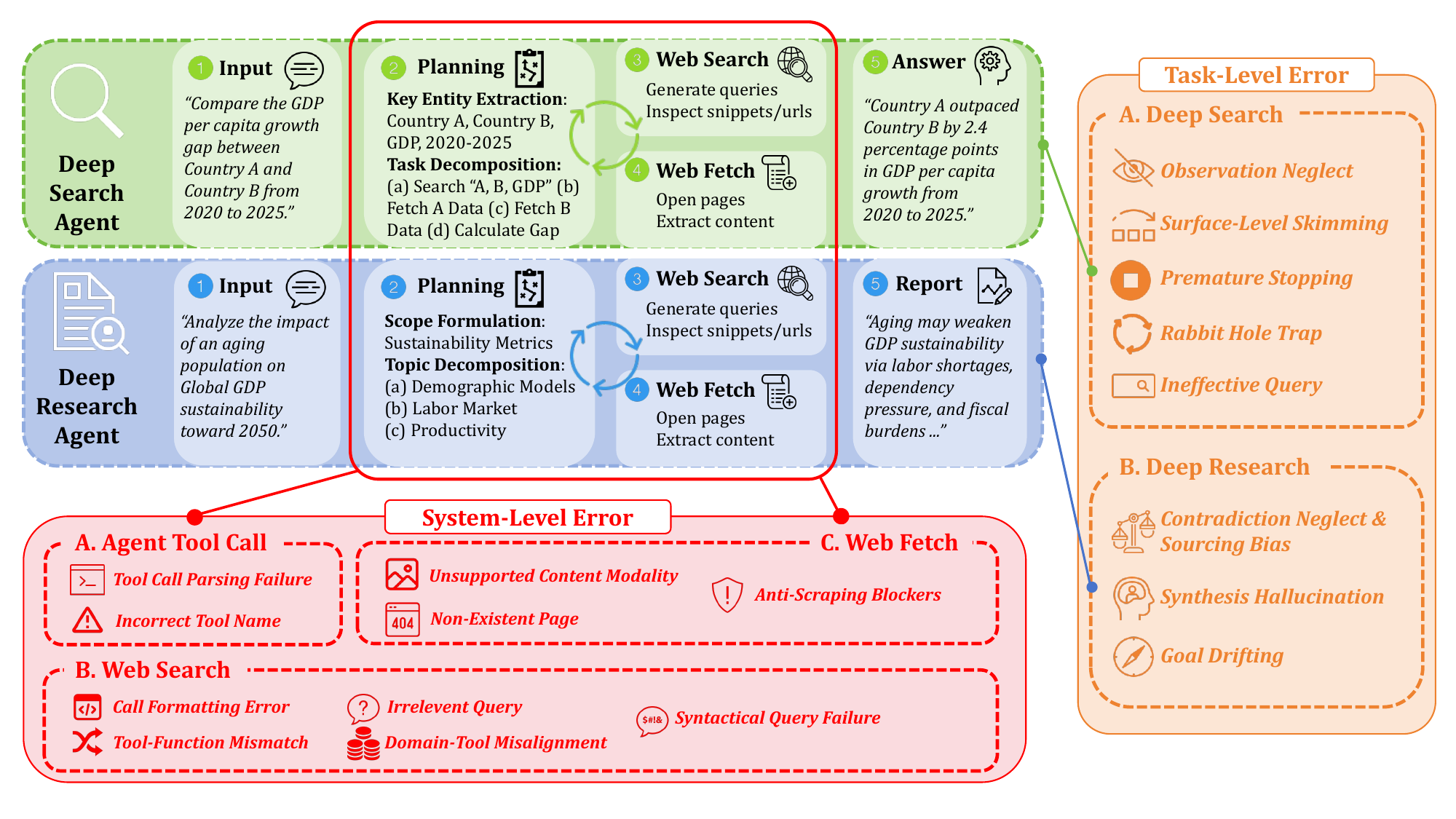} 
    \caption{\footnotesize Overview of Error Analysis of Deep Search \& Deep Research}
\end{figure*}

\section{Preliminary}
We analyze the process-level behaviors and architectural failure modes specific to Deep Search and Deep Research workflows. While Deep Search relies on iterative, goal-oriented exploration to pinpoint specific information across dynamic environments, Deep Research prioritizes the semantic synthesis of multi-source data into long-form analytical reports. This section formalizes these two paradigms and introduces the core terminology of our study.

\subsection{Deep Search Agents}

\textbf{Definition.} A Deep Search agent is an interactive decision-making system that incrementally acquires external evidence to answer a query requiring multi-step exploration. Unlike single-shot retrieval, the agent must repeatedly decide \emph{what to search next}, \emph{which source to inspect}, and \emph{when to terminate}.

\textbf{Formalization.} We model Deep Search as a finite-horizon interactive process with a minimal tool set
\[
\mathcal{A}_{\text{search}}=\{\texttt{web\_search},\texttt{web\_fetch},\texttt{stop}\}.
\]
Given query $q$, at step $t$ the agent maintains history $h_t=(q,o_{1:t},a_{1:t-1})$ and chooses an action. For \texttt{web\_search}, the agent submits a search query $u_t$ to a search engine and receives the top-$k$ ranked results
\[
R_t=\{(\text{title}_i,\text{url}_i,\text{snippet}_i)\}_{i=1}^{k},
\]
where each item contains title, URL, and snippet. For \texttt{web\_fetch}, the agent submits a URL and receives the corresponding raw webpage content $c(\text{url})$ (e.g., HTML or parsed text). The agent iterates between search and fetch until termination, then outputs an answer $\hat{y}$ with evidence set $E_T$.

\subsection{Deep Research Agents}

\textbf{Definition.} A Deep Research agent is a long-context synthesis system that transforms web-scale evidence into a structured analytical report. Compared with Deep Search, its primary challenge is not sparse fact lookup, but cross-source reconciliation, hierarchical organization, and logically consistent synthesis.

\textbf{Formalization.} We adopt the same minimal tool set as Deep Search,
\[
\mathcal{A}_{\text{research}}=\{\texttt{web\_search},\texttt{web\_fetch},\texttt{stop}\}.
\]
Given objective $q$, the agent iteratively invokes \texttt{web\_search} and \texttt{web\_fetch} to build an evidence pool $\mathcal{D}=\{d_1,\dots,d_n\}$, where search returns top-$k$ metadata triplets
\[
R_t=\{(\text{title}_i,\text{url}_i,\text{snippet}_i)\}_{i=1}^{k},
\]
and fetch returns raw page content $c(\text{url})$. It then generates a report
\[
 r = f_\theta(q,\mathcal{D}).
\]
We evaluate report quality with three coupled objectives: factual grounding $G(r,\mathcal{D})$, cross-source consistency $C(r,\mathcal{D})$, and task completeness $U(r,q)$. A high-quality report requires all three to be jointly satisfied rather than optimized in isolation.

\section{Methodology}
\label{Methodology}
We evaluate agent failures through a bifurcated methodology that explicitly categorizes breakdowns into System-Level errors and Task-Level errors. Because LLM agents operate at the interface of programmatic tool execution and probabilistic reasoning, this separation is essential to avoid conflating deterministic infrastructure faults with semantic cognitive limitations. Accordingly, our diagnostic pipeline combines rule-based parsing to isolate System-Level errors with a human-guided LLM-as-a-Judge framework to evaluate Task-Level errors.

\subsection{System-Level Errors}
To systematically identify system-level errors, we develop a hybrid diagnostic pipeline that combines rule-based extraction with LLM-assisted evaluation. First, we employ regular expressions and structured parsing to precisely extract the specific tool invocation and its corresponding reasoning at each step of the agent's trajectory. Once these action-thought pairs are isolated from the raw logs, they are submitted to an LLM evaluator. The LLM acts as a targeted judge, analyzing the semantic alignment between the agent's stated intent and the executed tool syntax to accurately diagnose system-level anomalies, such as tool-function mismatches or domain misalignments. The pipeline transitions from deterministic parsing directly to semantic validation. By having the parser handle the structural extraction, the LLM is freed from parsing raw logs and can focus purely on evaluating the logical correspondence between what the agent wanted to do and the tool it actually used.

We categorize these System-Level errors across three critical phases of the Deep Search and Deep Research pipeline: \textbf{Agent Tool Call},  \textbf{Web Search}, and \textbf{Web Fetch}.

\textbf{Agent Tool Call Errors.} This category encompasses failures where the LLM's raw generation cannot be translated into an actionable command by the environment parsing engine. We categorize these failures into two modes: \textbf{Tool Call Parsing Failure.} The agent fails to generate well-formed structural tags, preventing the system from extracting the intended action. \textbf{Incorrect Tool Name.} The agent hallucinates or incorrectly formats the name of a tool, invoking an API that does not exist in the predefined environment toolkit.

\textbf{Web Search Errors.} These errors occur when the agent successfully invokes a tool, but the provided arguments or query logic are fundamentally flawed, leading to immediate API rejections or guaranteed zero-hit returns. We categorize these failures into five modes: \textbf{Call Formatting Error.} The tool is invoked, but mandatory arguments are missing or violate the required JSON/string schema. \textbf{Tool-Function Mismatch.} The agent provides an argument that conflicts with the tool's intended utility. A common pattern identified in our logs is passing a raw URL into a text-based search engine instead of using a dedicated web-browsing tool.\textbf{Irrelevant Query.} The agent mistakenly inputs internal thoughts or system instructions as search queries. \textbf{Domain Tool Misalignment.} The agent uses an inappropriate specialized tool for a specific domain entity. For example, the agent identifies an arXiv preprint ID but erroneously queries it using a Wikipedia search tool, resulting in empty returns. \textbf{Syntactical Query Failure.} The generated query contains excessive special characters, regex artifacts, or bizarre string structures that overwhelm the search engine and inevitably yield zero results.

\textbf{Web-Fetch Retrieval Errors.} These errors are environment driven failures triggered during the actual execution of web scraping or document retrieval. We categorize these failures into three modes: \textbf{Unsupported Content Modality.} The agent attempts to read media types (e.g., audio, video, or highly complex multimodal formats) that the text-based web-scraping parser cannot process. \textbf{Non-Existent Page.} The agent attempts to scrape a webpage that has been removed, resulting in dead links or unresolved domains. \textbf{Anti-Scraping Blockers.} The fetch action is intercepted by server-side security mechanisms. We identify these by matching log keywords for HTTP 403 Forbidden, CAPTCHA challenges, or specific firewall intercepts, which prevent content extraction.

In summary, by categorizing System-Level errors across the agent tool call, web search and web fetch, we establish a structured taxonomy of deterministic failures. This classification ensures that these syntax and environment-driven breakdowns are explicitly recognized within the trajectory logs, preventing them from being misattributed as cognitive reasoning deficiencies during the Task-Level evaluation.

\subsection{Task-Level Errors}
When infrastructure execution succeeds, agents may still fail at the task level due to higher-order cognitive, strategic, or synthesis limitations that emerge across extended reasoning trajectories and large context windows. Because these semantic breakdowns are difficult to capture through rigid rule-based parsing alone, we adopt a human-guided LLM-as-a-Judge pipeline to systematically interpret the execution logs. To ensure our evaluation rubric is empirically grounded rather than arbitrarily defined, we first conduct a qualitative, human-in-the-loop analysis of sampled intermediate traces to inductively construct a behavioral taxonomy. Building upon this foundation, we then employ the LLM-as-a-Judge framework to automatically categorize these process-level deviations at scale. Given the distinct characteristics, we separately address the specific cognitive bottlenecks observed in Deep Search and Deep Research tasks, with the full judge prompts provided in Appendix~\ref{app:prompts}.

\textbf{Deep Search Error Taxonomy.} Deep Search tasks require agents to navigate unstructured environments to locate specific and often deeply embedded factual information. Task-level errors in this setting primarily manifest as failures in retrieval strategy and localized reasoning. We categorize these failures into five modes: \textbf{Observation Neglect.} The agent successfully retrieves relevant information within its environment observations but fails to extract or incorporate this evidence into subsequent reasoning steps. \textbf{Surface-Level Skimming.} The agent relies predominantly on short search-engine snippets rather than accessing underlying sources. When the required information is absent from snippets, the agent avoids deeper inspection and instead repeatedly reformulates queries or terminates exploration prematurely. \textbf{Premature Stopping.} The agent concludes the search process and produces a final answer before collecting all necessary evidence required to satisfy the task objective. This behavior frequently appears in exhaustive retrieval or multi-hop counting scenarios. \textbf{Rabbit Hole Trap.} The agent exhibits inefficient navigation by spending multiple consecutive steps traversing low-relevance sub-domains or entering repetitive navigation loops, without demonstrating effective backtracking or trajectory correction. \textbf{Ineffective Query.} The agent generates search queries that are overly broad, weakly grounded in task requirements, or semantically misaligned with the information need. In many cases, the agent repeatedly issues identical or minimally modified queries despite consistently uninformative results, indicating limited adaptive retrieval behavior.

\textbf{Deep Research Error Taxonomy.} Unlike targeted search tasks, Deep Research tasks require agents to autonomously aggregate, cross-reference, and synthesize large volumes of information into structured reports over extended interaction horizons. Task-level errors in this setting primarily arise from breakdowns in global context management and evidence synthesis. We categorize these failures into three modes: \textbf{Contradiction Neglect and Sourcing Bias.} Real-world information retrieval frequently produces conflicting evidence across sources. Agents may fail to reconcile such discrepancies through cross-validation, instead disproportionately relying on early or prominently encountered sources, leading to unverified claims propagated in the final report. \textbf{Synthesis Hallucination.} When available evidence is incomplete or insufficient, the agent generates synthesized content that is weakly grounded in retrieved sources. This commonly occurs when fulfilling complex structural requirements, where missing information is implicitly filled using parametric knowledge rather than verified observations. \textbf{Goal Drifting.} As the context window expands during long-horizon research trajectories, accumulated observations may dilute task salience. The agent increasingly attends to tangential details within retrieved content and ultimately produces outputs that deviate from key constraints or formatting requirements specified in the original user instruction.

\begin{table*}[t]
\centering
\small
\setlength{\tabcolsep}{5pt}
\renewcommand{\arraystretch}{1.12}

\begin{tabularx}{\textwidth}{
  >{\centering\arraybackslash}p{0.08\textwidth}
  >{\centering\arraybackslash}p{0.12\textwidth}
  >{\arraybackslash}m{0.25\textwidth}
  >{\arraybackslash}m{0.47\textwidth}
}
\toprule

\rowcolor{headerblue}
\textbf{Error Level}
& \textbf{Category}
& \textbf{Error Type}
& \textbf{Example} \\

\midrule


\multirow{16}{0.08\textwidth}{\centering System \\ Level}
& \multirow{2}{0.12\textwidth}{\centering Tool Call}
& \cellcolor{systemdark}
  Tool Call Parsing Failure
& \cellcolor{systemdark}
  The agent outputs a malformed tool call that cannot be parsed. \\

&
& \cellcolor{systemlight}
  Incorrect Tool Name
& \cellcolor{systemlight}
  The agent calls a non-existent tool such as \texttt{browse}. \\

\cmidrule(lr){2-4}


& \multirow{9}{0.12\textwidth}{\centering Web Search}
& \cellcolor{systemdark}
  Call Formatting Error
& \cellcolor{systemdark}
  The query is passed with invalid arguments or missing required fields. \\

&
& \cellcolor{systemlight}
  Tool--Function Mismatch
& \cellcolor{systemlight}
  The agent uses a search tool for a task that requires fetching a specific page. \\

&
& \cellcolor{systemdark}
  Irrelevant Query
& \cellcolor{systemdark}
  The generated query drifts away from the current sub-goal. \\

&
& \cellcolor{systemlight}
  Domain Tool Misalignment
& \cellcolor{systemlight}
  The agent searches the general web when a domain-specific source is required. \\

&
& \cellcolor{systemdark}
  Syntactical Query Failure
& \cellcolor{systemdark}
  The query contains broken operators, unmatched quotes, or malformed constraints. \\

\cmidrule(lr){2-4}


& \multirow{5}{0.12\textwidth}{\centering Web Fetch}
& \cellcolor{systemlight}
  Unsupported Content Modality
& \cellcolor{systemlight}
  The returned page requires video, image, or interactive content that the agent cannot access. \\

&
& \cellcolor{systemdark}
  Non-Existent Page
& \cellcolor{systemdark}
  The agent attempts to open a dead link or a 404 page. \\

&
& \cellcolor{systemlight}
  Anti-Scraping Blockers
& \cellcolor{systemlight}
  The page blocks automated access through CAPTCHA or anti-bot mechanisms. \\

\midrule


\multirow{13}{0.08\textwidth}{\centering Task \\ Level}
& \multirow{7}{0.12\textwidth}{\centering Deep Search}
& \cellcolor{taskdark}
  Observation Neglect
& \cellcolor{taskdark}
  The agent ignores useful evidence already present in retrieved snippets. \\

&
& \cellcolor{tasklight}
  Surface-Level Skimming
& \cellcolor{tasklight}
  The agent relies on snippets without opening or verifying the source page. \\

&
& \cellcolor{taskdark}
  Premature Stopping
& \cellcolor{taskdark}
  The agent stops searching before collecting sufficient evidence. \\

&
& \cellcolor{tasklight}
  Rabbit Hole Trap
& \cellcolor{tasklight}
  The agent repeatedly follows marginally related information paths. \\

&
& \cellcolor{taskdark}
  Ineffective Query
& \cellcolor{taskdark}
  The agent keeps issuing vague or poorly targeted queries. \\

\cmidrule(lr){2-4}


& \multirow{6}{0.12\textwidth}{\centering Deep Research}
& \cellcolor{tasklight}
  Contradiction Neglect \& Sourcing Bias
& \cellcolor{tasklight}
  The agent overlooks conflicting sources or over-relies on one source. \\

&
& \cellcolor{taskdark}
  Synthesis Hallucination
& \cellcolor{taskdark}
  The final report introduces claims not supported by retrieved evidence. \\

&
& \cellcolor{tasklight}
  Goal Drifting
& \cellcolor{tasklight}
  The generated analysis gradually shifts away from the original research objective. \\

\bottomrule
\end{tabularx}

\vspace{3pt}
\caption{Overview of the proposed error taxonomy across system-level and task-level failures, with representative examples for each error type.}
\label{tab:error_taxonomy}

\end{table*}

\section{Experiment}
\textbf{Setup}. To empirically validate our proposed error taxonomy and systematically diagnose bottlenecks in long-horizon autonomous agents, we design a controlled evaluation pipeline spanning both Deep Search and Deep Research paradigms. Importantly, our objective is not to benchmark end-to-end performance, but to construct experimental environments that expose process-level behaviors and failure patterns during extended agent execution. Each setup therefore emphasizes controllability and comparability, enabling consistent analysis of error emergence across different agent architectures and model backbones.

\textit{Deep Search.}
For Deep Search tasks, we employ the \textbf{Miroflow} \cite{su2026miroflow} agent framework and deliberately restrict the agent’s action space to core information-seeking operations. Specifically, the agent is equipped with only two tools: \texttt{search}, which retrieves search engine results and summaries, and \texttt{fetch}, which extracts full textual content from selected webpages. This constrained design minimizes confounding factors introduced by complex tool ecosystems and allows errors to be attributed primarily to navigation, retrieval, and evidence utilization behaviors. Such a controlled environment enables fine-grained observation of process-level failures, including \textbf{Surface-Level Skimming} and \textbf{Observation Neglect}. The framework is instantiated with two base models, \textbf{GPT-5-mini} \cite{singh2025openai} and \textbf{Mirothinker-30b} \cite{team2025mirothinker}, allowing us to examine whether observed error patterns remain consistent across different model families under identical interaction constraints.

\textit{Deep Research.}
For Deep Research tasks, which involve long-context reasoning and multi-source synthesis, we evaluate two research-oriented agent frameworks: \textbf{open-deep-research} and \textbf{step-deep-research}. These frameworks support iterative planning, large-context aggregation, and structured report generation over extended trajectories, reflecting realistic deep research workflows. Rather than optimizing task success rates, our configuration aims to capture how errors accumulate and propagate during prolonged reasoning and synthesis processes.The \textbf{open-deep-research} framework is implemented with \textbf{GPT-5}, while \textbf{step-deep-research} employs \textbf{Step-3.5-Flash}. This enables us to analyze whether error taxonomies generalize across distinct agent designs and execution dynamics.

\textbf{Benchmarks and Datasets.}
\label{sec:datasets}
To ground our error analysis in realistic agent workflows, we draw tasks from a collection of established and widely adopted benchmarks spanning both agentic paradigms. Rather than using these benchmarks for conventional performance evaluation, we leverage them as representative task sources to elicit diverse long-horizon agent trajectories suitable for diagnostic analysis. Detailed dataset statistics and configurations are provided in Appendix~\ref{app:dataset_details}. For Deep Search agent, to capture representative information-seeking and web navigation behaviors, we sample tasks from four commonly used Deep Search benchmarks: \textbf{BrowseComp \cite{wei2025browsecomp}} A benchmark frequently used to study complex web browsing and autonomous navigation across heterogeneous HTML environments. \textbf{BrowseComp-zh \cite{zhou2025browsecomp}} A standardized Chinese-language counterpart emphasizing multilingual web navigation and retrieval scenarios. \textbf{GAIA-val-text-only \cite{mialon2023gaia}} A challenging benchmark for multi-step question answering and tool-assisted reasoning. We use the text-only validation split to focus specifically on textual retrieval and reading processes. \textbf{XBench-DS \cite{chen2025xbench}} A dataset designed around iterative query reformulation and multi-hop web retrieval, representative of typical deep search workflows. For Deep Research agent, to study long-horizon synthesis and analytical report generation, we draw tasks from \textbf{Deep Research Bench} \cite{du2025deepresearch}, a benchmark commonly used to evaluate autonomous research agents on open-ended analytical requests requiring multi-source aggregation and structured synthesis.

Although these benchmarks are traditionally employed for end-to-end accuracy evaluation, our framework instead uses them to generate complete execution traces. By leveraging their curated task prompts, reference answers, and structural requirements, we obtain standardized problem settings that enable consistent, process-oriented error diagnosis across diverse agent trajectories. We adopt GPT-5 as the LLM judge for both system-level error and task-level error.

\textbf{Result.} In this section, we report process-level findings from both Deep Search and Deep Research settings under the unified taxonomy introduced in Section~\ref{Methodology}. We first analyze System-Level errors to characterize interaction-side reliability, and then analyze Task-Level errors to reveal paradigm-specific cognitive bottlenecks in retrieval and synthesis workflows. These results provide a fine-grained view of where current agent systems fail and which components should be prioritized for improvement.

\textit{System-Level Error}. We conduct an empirical analysis centered on Deep Search traces, with a primary focus on the system-level errors of \textbf{MiroThinker-30B}. This setting provides a clear basis for cross-dataset comparison under a unified execution pipeline.

Concretely, for each Deep Search benchmark, we report both trajectory-scale and error-composition statistics, including average steps per task, total executed steps, total system-level error steps, and category-wise counts of system-level failures. This supports normalized cross-dataset comparison by jointly characterizing overall interaction burden and the dominant failure profile under the same system-level taxonomy.

\begin{table*}[htbp]

\centering

\resizebox{2\columnwidth}{!}{%
\begin{tabular*}{0.9\textwidth}{@{}l c c c c@{}}
\toprule
\textbf{Metric} & \textbf{BrowseComp} & \textbf{BrowseComp-zh} & \textbf{GAIA-val-text-only} & \textbf{XBench-DS} \\

\midrule
\textbf{No Error  }& 3504 & 4861 & 1217 & 2750 \\

\midrule
Tool Call Parsing Failure & 22 & 6 & 3 & 15 \\
Incorrect Tool Name & 109 & 70 & 232& 236 \\

\midrule
Call Formatting Error & 4 & 13 & 34 & 23 \\
Tool-Function Mismatch & 434 & 286 & 1227 & 650 \\
Irrelevant Query & 30 & 36 & 169 & 22 \\
Domain Tool Misalignment & 72 & 426 & 1163 & 369 \\
Syntactical Query Failure\textbf{ }& 84 & 6 & 180 & 15 \\

\midrule
Unsupported Content Modality & 5 & 8 & 61 & 13 \\
Non-Existent Page & 0 & 1 & 39 & 2 \\
Anti-Scraping & 392 & 437 & 30 & 268 \\

\midrule

New Pattern Errors  & 10 & 20 & 7 & 13 \\

\midrule
\textbf{Avg. Steps / Task} & 46.66 & 61.7 & 42.34  & 43.76 \\
\textbf{Total Steps } & 4666 & 6170 & 4362 & 4376 \\
\textbf{Error Steps} & 1162 & 1309 & 3145 & 1626 \\
\midrule
\textbf{No Error Rate}  & 75.10 & 78.78 & 27.90& 62.84 \\
\textbf{Acc} & 17.00 & 42.00 & 63.11 & 73.00 \\
\bottomrule
\end{tabular*}%
}
\vspace{3pt}
\caption{System-Level error summary for Deep Search traces.}
\label{tab:system_level_error_summary}

\end{table*}

From Table \ref{tab:system_level_error_summary}, we observe clear dataset-dependent differences in system-level robustness. XBench-DS exhibits the highest concentration of system-level failures, whereas BrowseComp and BrowseComp-zh remain comparatively stable. This suggests that interaction-layer difficulty is shaped not only by trajectory length, but also by how strongly a benchmark stresses tool invocation and environment interaction.

Error compositions further reveal dataset-specific bottlenecks. BrowseComp is primarily constrained by tool–function alignment issues and web accessibility limitations, highlighting the joint importance of action precision and environmental reliability. BrowseComp-zh shows stronger domain–tool alignment challenges alongside accessibility failures, indicating increased pressure on tool selection under multilingual or domain-shifted conditions. In contrast, XBench-DS is dominated by tool orchestration failures, where functional mismatches outweigh low-level parsing errors. Together, these results suggest that system-level failures are structurally heterogeneous across benchmarks.

More broadly, the analysis reveals a shift in the operational bottlenecks of autonomous web agents. Purely syntactic failures are now rare, indicating that modern language models generally handle the mechanics of tool invocation reliably. Instead, systemic friction arises mainly from semantic misalignment, such as inappropriate tool usage, weakly grounded queries, and resistance from real-world web environments. Notably, frequent system-level disruptions do not necessarily preclude successful task completion. Agents often recover through iterative adjustment and continued interaction, compensating for imperfect instruction adherence or environmental obstacles. This observation suggests that robustness in open environments emerges less from flawless execution than from an agent’s ability to detect, adapt to, and recover from inevitable failures during interaction.

\textit{Task-Level Error}. Task-level errors capture higher-order cognitive and strategic failures that occur even when tool execution is syntactically valid. Unlike system-level breakdowns, these errors are reflected in how agents explore, accumulate evidence, and decide when to stop. In this section, we analyze task-level error distributions across benchmarks and models to identify paradigm-specific bottlenecks in Deep Search and Deep Research workflows.

As detailed in Table \ref{tab:deep_search_errors}, our task-level evaluation of Deep Search agents reveals a strong positive correlation between process-level reliability (No Error Rate) and final task accuracy (Acc). Models that successfully navigate intermediate cognitive bottlenecks consistently achieve higher end-to-end success, confirming the diagnostic validity of our taxonomy. Furthermore, the error distributions expose stark behavioral divergences between model architectures.  GPT-5-mini predominantly fail due to Premature Stop and Surface Skim, which indicates an inherent reluctance to sustain multi-hop exploration. Conversely, the open-weight Miro-30B demonstrates strong execution endurance, but heavily bottlenecked by Ineffective Query formulation and Obs. Neglect. This highlights a critical gap in strategic reasoning, that while the agent is willing to search extensively, it struggles to translate complex intents into optimal search syntax or to extract target signals from dense retrieved contexts.
Table \ref{tab:deep_research_errors} demonstrates that different Deep Research frameworks succumb to entirely distinct cognitive vulnerabilities. The Open Deep Research framework maintains a relatively higher process reliability but is primarily sabotaged by Goal Drifting. As the context window expands during extensive evidence gathering, the framework frequently loses track of core user constraints and formatting instructions. In contrast, Step Deep Research struggles with Synthesis Hallucination rather than Goal Drifting, which indicate that it attempts to forcefully fulfill structural requirements by fabricating data when sufficient factual evidence is missing. Additionally, Step Deep Research exhibits a higher propensity for compounded errors, such as the dangerous combination of Contradiction Neglect and Sourcing Bias with Synth Hallucination.This highlight how aggressive synthesis protocols can trigger cascading cognitive failures if the underlying retrieval phase is fundamentally flawed.

\begin{table}[t]
\centering
\small
\setlength{\tabcolsep}{3.5pt}
\renewcommand{\arraystretch}{1.05}
\caption{Distribution of task-level errors in Deep Search agents. Miro-30B denotes MiroThinker-30B-v1.5.}
\label{tab:deep_search_errors}
\resizebox{\columnwidth}{!}{%
\begin{tabular}{@{} l cc c cc @{}} 
\toprule
& \multicolumn{2}{c}{\textbf{BrowseComp}} && \multicolumn{2}{c}{\textbf{BrowseComp-zh}} \\
\cmidrule{2-3} \cmidrule{5-6}
\textbf{Failure Mode} & \textbf{Miro-30B} & \textbf{GPT-5-mini} && \textbf{Miro-30B} & \textbf{GPT-5-mini} \\
\midrule
No Error              & 17 & 14 && 45 & 27 \\
Obs. Neglect          & 12 & 2  && 8  & 1  \\
Surface Skim.         & 17 & 20 && 9  & 14 \\
Premature Stop        & 4  & 20 && 1  & 38 \\
Rabbit Hole           & 9  & 3  && 8  & 0  \\
Ineffective Query     & 47 & 41 && 25 & 20 \\
\midrule
\textbf{No Error Rate} & \textbf{17.00} & \textbf{14.00} && \textbf{45.00} & \textbf{27.00} \\
\textbf{Acc}           & \textbf{17.00} & \textbf{15.00} && \textbf{42.00} & \textbf{29.00} \\
\midrule
\midrule
& \multicolumn{2}{c}{\textbf{GAIA}} && \multicolumn{2}{c}{\textbf{XBench-DS}} \\
\cmidrule{2-3} \cmidrule{5-6}
\textbf{Failure Mode} & \textbf{Miro-30B} & \textbf{GPT-5-mini} && \textbf{Miro-30B} & \textbf{GPT-5-mini} \\
\midrule
No Error              & 64 & 53 && 72 & 19 \\
Obs. Neglect          & 12 & 1  && 18 & 2  \\
Surface Skim.         & 5  & 4  && 3  & 0  \\
Premature Stop        & 12 & 41 && 1  & 75 \\
Rabbit Hole           & 7  & 1  && 2  & 1  \\
Ineffective Query     & 3  & 3  && 4  & 3  \\
\midrule
\textbf{No Error Rate} & \textbf{62.14} & \textbf{51.46} && \textbf{72.00} & \textbf{19.00} \\
\textbf{Acc}           & \textbf{63.11} & \textbf{56.31} && \textbf{73.00} & \textbf{41.00} \\
\bottomrule
\end{tabular}%
}
\end{table}

\begin{table}[t]
\centering
\footnotesize
\setlength{\tabcolsep}{4pt}
\renewcommand{\arraystretch}{0.94}
\caption{Distribution of task-level errors in Deep Research agents. C denotes Contradiction Neglect and Sourcing Bias, S denotes Synthesis Hallucination, and G denotes Goal Drift.}
\label{tab:deep_research_errors}
\resizebox{\columnwidth}{!}{%
\begin{tabular}{@{} l cc @{}} 
\toprule
\textbf{Failure Mode} & \textbf{Open-DR} & \textbf{Step-DR} \\
\midrule
No Error & 44 & 13 \\
\midrule
Contradiction Neglecet \& Sourcing Bias (C) & 10 & 23 \\
Synthesis Hallucination (S)      & 25 & 80 \\
Goal Drifting (G)       & 51 & 7  \\
\midrule
C only       & 0  & 3  \\
S only       & 5  & 60 \\
G only       & 30 & 2  \\
C + S        & 0  & 17 \\
S + G        & 11 & 2  \\
C + G        & 1  & 2  \\
C + S + G    & 9  & 1  \\
\midrule
\textbf{No Error Rate} & \textbf{44} & 13 \\
\textbf{RACE}          & \textbf{0.536} & 0.4802 \\
\textbf{FACT}          & \textbf{0.468} & 0.4143 \\
\bottomrule
\end{tabular}%
}
\end{table}
\section{Discussion}

Our fine-grained, trace-level evaluation provides actionable insights for designing next-generation Deep Search and Deep Research agents. Unlike end-to-end metrics that only assess final correctness, our framework identifies where and how failures emerge throughout the operational trajectory. This process-level visibility is particularly important for long-horizon agents, where small intermediate deviations—such as misaligned queries or overlooked evidence—can accumulate into downstream failure. By exposing these hidden bottlenecks, our taxonomy connects empirical evaluation with concrete system design considerations.

\textbf{Implications for Model Training.}
The separation between System-Level and Task-Level errors enables more targeted supervision strategies. Current training pipelines often penalize deterministic execution mistakes and genuine reasoning failures equally, despite their fundamentally different causes. Our taxonomy suggests decoupling these objectives: models can first learn reliable tool execution and formatting behaviors, while higher-level training focuses on exploration strategy, evidence integration, and synthesis quality. This perspective naturally supports trace-aware optimization objectives that reward intermediate reasoning quality, evidence coverage, and recovery after failed actions rather than final-answer accuracy alone. Moreover, the taxonomy provides structured reward signals for iterative fine-tuning and Reinforcement Learning from AI Feedback (RLAIF), allowing training procedures to discourage trajectory-level behaviors such as premature termination or inefficient exploration.

\textbf{Implications for Toolset Design.}
System-level error patterns indicate that tools should function not merely as passive interfaces but as stabilizing components within the agent–environment loop. Improved schema constraints, informative semantic error feedback, and built-in recovery affordances can substantially reduce cascading failures. For Deep Search, tools should support progressive evidence accumulation and abstract away low-level navigation burdens such as pagination handling. For Deep Research, where long-context overload becomes dominant, tools should facilitate source alignment, contradiction detection, and citation-grounded synthesis. In this sense, tool design becomes a central factor shaping agent robustness rather than a peripheral implementation detail.

\textbf{Implications for Agentic Frameworks.}
Observed error propagation further highlights limitations of monolithic reasoning loops. When planning, navigation, reading, and synthesis are handled within a single continuous process, context degradation and strategic drift become inevitable. More robust frameworks should introduce modular control stages separating planning, verification, and repair. Deep Search agents benefit from adaptive stopping criteria and explicit exploration-coverage checks, while Deep Research systems require staged synthesis pipelines that isolate retrieval from reconciliation and report generation. Incorporating lightweight runtime self-diagnosis mechanisms based on our taxonomy can enable early intervention, allowing agents to revise trajectories before failures fully propagate.

\textbf{Broader Takeaway.}
Fine-grained evaluation serves as an enabling infrastructure for reliable autonomous agents. By transforming opaque failures into structured diagnostic signals, it supports systematic improvements across training, tool ecosystems, and framework architecture. As agent autonomy and interaction horizons continue to expand, shifting from outcome-based evaluation toward process-level diagnosis is essential for developing systems that are not only capable, but consistently reliable in open environments.

\section{Conclusion}
In this work, we present a unified, process-oriented framework for evaluating Deep Search and Deep Research agents through full execution traces rather than final outcomes alone. By separating failures into System-Level and Task-Level categories, our framework makes intermediate failure mechanisms observable and comparable across paradigms. Our empirical findings show that performance bottlenecks often stem from error propagation during interaction, rather than solely from weaknesses in final response generation, enabling a more faithful diagnosis of agent reliability in long-horizon workflows. Beyond evaluation, the proposed taxonomy offers practical guidance for agent design, supporting targeted training objectives, more robust tool interfaces, and modular architectures with explicit planning, verification, and repair mechanisms. We hope this work contributes toward agents that are not only accurate on benchmarks but also process-stable in realistic deployments. Future work will extend this framework to broader environments, richer tool ecosystems, and automated online error recovery.




\bibliography{example_paper}

@inproceedings{yao2022react,
  title={React: Synergizing reasoning and acting in language models},
  author={Yao, Shunyu and Zhao, Jeffrey and Yu, Dian and Du, Nan and Shafran, Izhak and Narasimhan, Karthik R and Cao, Yuan},
  booktitle={The eleventh international conference on learning representations},
  year={2022}
}

@inproceedings{wang2023plan,
  title={Plan-and-solve prompting: Improving zero-shot chain-of-thought reasoning by large language models},
  author={Wang, Lei and Xu, Wanyu and Lan, Yihuai and Hu, Zhiqiang and Lan, Yunshi and Lee, Roy Ka-Wei and Lim, Ee-Peng},
  booktitle={Proceedings of the 61st annual meeting of the association for computational linguistics (volume 1: long papers)},
  pages={2609--2634},
  year={2023}
}

@article{guo2024large,
  title={Large language model based multi-agents: A survey of progress and challenges},
  author={Guo, Taicheng and Chen, Xiuying and Wang, Yaqi and Chang, Ruidi and Pei, Shichao and Chawla, Nitesh V and Wiest, Olaf and Zhang, Xiangliang},
  journal={arXiv preprint arXiv:2402.01680},
  year={2024}
}

@article{wei2025browsecomp,
  title={Browsecomp: A simple yet challenging benchmark for browsing agents},
  author={Wei, Jason and Sun, Zhiqing and Papay, Spencer and McKinney, Scott and Han, Jeffrey and Fulford, Isa and Chung, Hyung Won and Passos, Alex Tachard and Fedus, William and Glaese, Amelia},
  journal={arXiv preprint arXiv:2504.12516},
  year={2025}
}

@inproceedings{mialon2023gaia,
  title={Gaia: a benchmark for general ai assistants},
  author={Mialon, Gr{\'e}goire and Fourrier, Cl{\'e}mentine and Wolf, Thomas and LeCun, Yann and Scialom, Thomas},
  booktitle={The Twelfth International Conference on Learning Representations},
  year={2023}
}

@article{luo2025large,
  title={Large language model agent: A survey on methodology, applications and challenges},
  author={Luo, Junyu and Zhang, Weizhi and Yuan, Ye and Zhao, Yusheng and Yang, Junwei and Gu, Yiyang and Wu, Bohan and Chen, Binqi and Qiao, Ziyue and Long, Qingqing and others},
  journal={arXiv preprint arXiv:2503.21460},
  year={2025}
}

@article{xi2025survey,
  title={A survey of llm-based deep search agents: Paradigm, optimization, evaluation, and challenges},
  author={Xi, Yunjia and Lin, Jianghao and Xiao, Yongzhao and Zhou, Zheli and Shan, Rong and Gao, Te and Zhu, Jiachen and Liu, Weiwen and Yu, Yong and Zhang, Weinan},
  journal={arXiv preprint arXiv:2508.05668},
  year={2025}
}

@article{xu2025comprehensive,
  title={A comprehensive survey of deep research: Systems, methodologies, and applications},
  author={Xu, Renjun and Peng, Jingwen},
  journal={arXiv preprint arXiv:2506.12594},
  year={2025}
}

@article{schick2023toolformer,
  title={Toolformer: Language models can teach themselves to use tools},
  author={Schick, Timo and Dwivedi-Yu, Jane and Dess{\`\i}, Roberto and Raileanu, Roberta and Lomeli, Maria and Hambro, Eric and Zettlemoyer, Luke and Cancedda, Nicola and Scialom, Thomas},
  journal={Advances in neural information processing systems},
  volume={36},
  pages={68539--68551},
  year={2023}
}

@article{achiam2023gpt,
  title={Gpt-4 technical report},
  author={Achiam, Josh and Adler, Steven and Agarwal, Sandhini and Ahmad, Lama and Akkaya, Ilge and Aleman, Florencia Leoni and Almeida, Diogo and Altenschmidt, Janko and Altman, Sam and Anadkat, Shyamal and others},
  journal={arXiv preprint arXiv:2303.08774},
  year={2023}
}

@article{li2025websailor,
  title={Websailor: Navigating super-human reasoning for web agent},
  author={Li, Kuan and Zhang, Zhongwang and Yin, Huifeng and Zhang, Liwen and Ou, Litu and Wu, Jialong and Yin, Wenbiao and Li, Baixuan and Tao, Zhengwei and Wang, Xinyu and others},
  journal={arXiv preprint arXiv:2507.02592},
  year={2025}
}

@article{zhou2023webarena,
  title={Webarena: A realistic web environment for building autonomous agents},
  author={Zhou, Shuyan and Xu, Frank F and Zhu, Hao and Zhou, Xuhui and Lo, Robert and Sridhar, Abishek and Cheng, Xianyi and Ou, Tianyue and Bisk, Yonatan and Fried, Daniel and others},
  journal={arXiv preprint arXiv:2307.13854},
  year={2023}
}

@article{yang2025qwen3,
  title={Qwen3 technical report},
  author={Yang, An and Li, Anfeng and Yang, Baosong and Zhang, Beichen and Hui, Binyuan and Zheng, Bo and Yu, Bowen and Gao, Chang and Huang, Chengen and Lv, Chenxu and others},
  journal={arXiv preprint arXiv:2505.09388},
  year={2025}
}

@article{zhu2025llm,
  title={Where llm agents fail and how they can learn from failures},
  author={Zhu, Kunlun and Liu, Zijia and Li, Bingxuan and Tian, Muxin and Yang, Yingxuan and Zhang, Jiaxun and Han, Pengrui and Xie, Qipeng and Cui, Fuyang and Zhang, Weijia and others},
  journal={arXiv preprint arXiv:2509.25370},
  year={2025}
}

@inproceedings{mohammadi2025evaluation,
  title={Evaluation and benchmarking of llm agents: A survey},
  author={Mohammadi, Mahmoud and Li, Yipeng and Lo, Jane and Yip, Wendy},
  booktitle={Proceedings of the 31st ACM SIGKDD Conference on Knowledge Discovery and Data Mining V. 2},
  pages={6129--6139},
  year={2025}
}

@article{du2025deepresearch,
  title={Deepresearch bench: A comprehensive benchmark for deep research agents},
  author={Du, Mingxuan and Xu, Benfeng and Zhu, Chiwei and Wang, Xiaorui and Mao, Zhendong},
  journal={arXiv preprint arXiv:2506.11763},
  year={2025}
}

@article{nakano2021webgpt,
  title={Webgpt: Browser-assisted question-answering with human feedback},
  author={Nakano, Reiichiro and Hilton, Jacob and Balaji, Suchir and Wu, Jeff and Ouyang, Long and Kim, Christina and Hesse, Christopher and Jain, Shantanu and Kosaraju, Vineet and Saunders, William and others},
  journal={arXiv preprint arXiv:2112.09332},
  year={2021}
}

@inproceedings{liu2023webglm,
  title={WebGLM: towards an efficient web-enhanced question answering system with human preferences},
  author={Liu, Xiao and Lai, Hanyu and Yu, Hao and Xu, Yifan and Zeng, Aohan and Du, Zhengxiao and Zhang, Peng and Dong, Yuxiao and Tang, Jie},
  booktitle={Proceedings of the 29th ACM SIGKDD conference on knowledge discovery and data mining},
  pages={4549--4560},
  year={2023}
}

@article{shinn2023reflexion,
  title={Reflexion: Language agents with verbal reinforcement learning},
  author={Shinn, Noah and Cassano, Federico and Gopinath, Ashwin and Narasimhan, Karthik and Yao, Shunyu},
  journal={Advances in neural information processing systems},
  volume={36},
  pages={8634--8652},
  year={2023}
}

@inproceedings{he2024webvoyager,
  title={Webvoyager: Building an end-to-end web agent with large multimodal models},
  author={He, Hongliang and Yao, Wenlin and Ma, Kaixin and Yu, Wenhao and Dai, Yong and Zhang, Hongming and Lan, Zhenzhong and Yu, Dong},
  booktitle={Proceedings of the 62nd Annual Meeting of the Association for Computational Linguistics (Volume 1: Long Papers)},
  pages={6864--6890},
  year={2024}
}

@inproceedings{shao2024assisting,
  title={Assisting in writing wikipedia-like articles from scratch with large language models},
  author={Shao, Yijia and Jiang, Yucheng and Kanell, Theodore and Xu, Peter and Khattab, Omar and Lam, Monica},
  booktitle={Proceedings of the 2024 Conference of the North American Chapter of the Association for Computational Linguistics: Human Language Technologies (Volume 1: Long Papers)},
  pages={6252--6278},
  year={2024}
}

@misc{openai2025deepresearch,
  author = {{OpenAI}},
  title = {Deep Research},
  year = {2025},
  howpublished = {\url{https://openai.com/index/introducing-deep-research/}},
}

@misc{google2026geminideepresearch,
  author = {{Google LLC}},
  title = {Gemini Deep Research},
  year = {2024},
  howpublished = {\url{https://gemini.google/overview/deep-research/}},
}

@inproceedings{yang2018hotpotqa,
  title={HotpotQA: A dataset for diverse, explainable multi-hop question answering},
  author={Yang, Zhilin and Qi, Peng and Zhang, Saizheng and Bengio, Yoshua and Cohen, William and Salakhutdinov, Ruslan and Manning, Christopher D},
  booktitle={Proceedings of the 2018 conference on empirical methods in natural language processing},
  pages={2369--2380},
  year={2018}
}

@article{deng2023mind2web,
  title={Mind2web: Towards a generalist agent for the web},
  author={Deng, Xiang and Gu, Yu and Zheng, Boyuan and Chen, Shijie and Stevens, Sam and Wang, Boshi and Sun, Huan and Su, Yu},
  journal={Advances in Neural Information Processing Systems},
  volume={36},
  pages={28091--28114},
  year={2023}
}

@article{ji2023survey,
  title={Survey of hallucination in natural language generation},
  author={Ji, Ziwei and Lee, Nayeon and Frieske, Rita and Yu, Tiezheng and Su, Dan and Xu, Yan and Ishii, Etsuko and Bang, Ye Jin and Madotto, Andrea and Fung, Pascale},
  journal={ACM computing surveys},
  volume={55},
  number={12},
  pages={1--38},
  year={2023},
  publisher={ACM New York, NY}
}

@article{kadavath2022language,
  title={Language models (mostly) know what they know},
  author={Kadavath, Saurav and Conerly, Tom and Askell, Amanda and Henighan, Tom and Drain, Dawn and Perez, Ethan and Schiefer, Nicholas and Hatfield-Dodds, Zac and DasSarma, Nova and Tran-Johnson, Eli and others},
  journal={arXiv preprint arXiv:2207.05221},
  year={2022}
}

@inproceedings{press2023measuring,
  title={Measuring and narrowing the compositionality gap in language models},
  author={Press, Ofir and Zhang, Muru and Min, Sewon and Schmidt, Ludwig and Smith, Noah A and Lewis, Mike},
  booktitle={Findings of the Association for Computational Linguistics: EMNLP 2023},
  pages={5687--5711},
  year={2023}
}

@article{gendron2023large,
  title={Large language models are not strong abstract reasoners},
  author={Gendron, Ga{\"e}l and Bao, Qiming and Witbrock, Michael and Dobbie, Gillian},
  journal={arXiv preprint arXiv:2305.19555},
  year={2023}
}

@article{zhou2025browsecomp,
  title={Browsecomp-zh: Benchmarking web browsing ability of large language models in chinese},
  author={Zhou, Peilin and Leon, Bruce and Ying, Xiang and Zhang, Can and Shao, Yifan and Ye, Qichen and Chong, Dading and Jin, Zhiling and Xie, Chenxuan and Cao, Meng and others},
  journal={arXiv preprint arXiv:2504.19314},
  year={2025}
}

@article{chen2025xbench,
  title={xbench: Tracking agents productivity scaling with profession-aligned real-world evaluations},
  author={Chen, Kaiyuan and Ren, Yixin and Liu, Yang and Hu, Xiaobo and Tian, Haotong and Xie, Tianbao and Liu, Fangfu and Zhang, Haoye and Liu, Hongzhang and Gong, Yuan and others},
  journal={arXiv preprint arXiv:2506.13651},
  year={2025}
}

@article{patil2024gorilla,
  title={Gorilla: Large language model connected with massive apis},
  author={Patil, Shishir G and Zhang, Tianjun and Wang, Xin and Gonzalez, Joseph E},
  journal={Advances in Neural Information Processing Systems},
  volume={37},
  pages={126544--126565},
  year={2024}
}

@article{qin2023toolllm,
  title={Toolllm: Facilitating large language models to master 16000+ real-world apis},
  author={Qin, Yujia and Liang, Shihao and Ye, Yining and Zhu, Kunlun and Yan, Lan and Lu, Yaxi and Lin, Yankai and Cong, Xin and Tang, Xiangru and Qian, Bill and others},
  journal={arXiv preprint arXiv:2307.16789},
  year={2023}
}

@article{lewis2020retrieval,
  title={Retrieval-augmented generation for knowledge-intensive nlp tasks},
  author={Lewis, Patrick and Perez, Ethan and Piktus, Aleksandra and Petroni, Fabio and Karpukhin, Vladimir and Goyal, Naman and K{\"u}ttler, Heinrich and Lewis, Mike and Yih, Wen-tau and Rockt{\"a}schel, Tim and others},
  journal={Advances in neural information processing systems},
  volume={33},
  pages={9459--9474},
  year={2020}
}

@inproceedings{wu2024autogen,
  title={Autogen: Enabling next-gen LLM applications via multi-agent conversations},
  author={Wu, Qingyun and Bansal, Gagan and Zhang, Jieyu and Wu, Yiran and Li, Beibin and Zhu, Erkang and Jiang, Li and Zhang, Xiaoyun and Zhang, Shaokun and Liu, Jiale and others},
  booktitle={First conference on language modeling},
  year={2024}
}

@article{su2026miroflow,
  title={MiroFlow: Towards High-Performance and Robust Open-Source Agent Framework for General Deep Research Tasks},
  author={Su, Shiqian and Xing, Sen and Dong, Xuan and Zhong, Muyan and Wang, Bin and Zhu, Xizhou and Chen, Yuntao and Wang, Wenhai and Deng, Yue and Zhu, Pengxiang and others},
  journal={arXiv preprint arXiv:2602.22808},
  year={2026}
}

@article{team2025mirothinker,
  title={Mirothinker: Pushing the performance boundaries of open-source research agents via model, context, and interactive scaling},
  author={Team, MiroMind and Bai, Song and Bing, Lidong and Chen, Carson and Chen, Guanzheng and Chen, Yuntao and Chen, Zhe and Chen, Ziyi and Dai, Jifeng and Dong, Xuan and others},
  journal={arXiv preprint arXiv:2511.11793},
  year={2025}
}

@article{singh2025openai,
  title={Openai gpt-5 system card},
  author={Singh, Aaditya and Fry, Adam and Perelman, Adam and Tart, Adam and Ganesh, Adi and El-Kishky, Ahmed and McLaughlin, Aidan and Low, Aiden and Ostrow, AJ and Ananthram, Akhila and others},
  journal={arXiv preprint arXiv:2601.03267},
  year={2025}
}
\bibliographystyle{icml2026}

\appendix

\section{Related Work}
\label{app:related_work}
\subsection{Evolution of Deep Search and Deep Research Agents}
The paradigm of Large Language Models (LLMs) has fundamentally shifted from static, closed-book text generators to dynamic, environment-grounded autonomous agents. Early efforts demonstrated that augmenting LLMs with external tools—such as search engines and computation APIs—significantly mitigates knowledge cut-off limitations and reduces factual hallucinations. Pioneering works like WebGPT \cite{nakano2021webgpt}, WebGLM \cite{liu2023webglm}, and Toolformer \cite{schick2023toolformer} established that language models could generate explicit API calls to offload factual retrieval to external environments. Subsequent developments in tool-augmented LLMs, such as Gorilla \cite{patil2024gorilla} and ToolLLM \cite{qin2023toolllm}, further expanded the versatility of models to interact with vast, real-world API ecosystems.

However, as user queries evolved from simple factual lookups to complex, knowledge-intensive tasks, standard single-turn Retrieval-Augmented Generation (RAG) \cite{lewis2020retrieval} proved fundamentally insufficient. A primary limitation of traditional RAG systems is their inherent reliance on static, pre-indexed local corpora, which renders them incapable of autonomously exploring and acquiring up-to-date knowledge from the open, dynamically evolving internet environment. This critical bottleneck catalyzed the emergence of the Deep Search paradigm.

Built upon interleaved reasoning and acting frameworks like ReAct \cite{yao2022react}, hierarchical planning mechanisms like Plan-and-Solve \cite{wang2023plan}, and self-correcting loops like Reflexion \cite{shinn2023reflexion}, Deep Search agents are engineered to execute iterative, multi-hop information retrieval. Rather than fetching a single snippet from a closed database, these agents actively engage with the web. Driven by the need to solve complex tasks evaluated in realistic benchmarks like WebArena \cite{zhou2023webarena} and Mind2Web \cite{deng2023mind2web}, Deep Search frameworks such as WebSailor \cite{li2025websailor} and WebVoyager \cite{he2024webvoyager} enable agents to autonomously navigate complex HTML Document Object Models (DOM), traverse pagination, filter irrelevant hyperlinks, and dynamically refine their search queries based on live environmental observations.

Concurrently, the frontier of agentic capabilities has expanded beyond mere information acquisition toward open-ended synthesis and long-horizon planning, giving rise to Deep Research architectures. While Deep Search focuses on the precise retrieval of targeted facts across multiple web hops, Deep Research demands that agents autonomously compile, evaluate, and synthesize massive volumes of unstructured information into comprehensive reports over extended temporal horizons. Proprietary commercial systems, such as OpenAI's Deep Research \cite{openai2025deepresearch} and Gemini Deep Research \cite{google2026geminideepresearch}, have recently demonstrated advanced synthesis engines capable of orchestrating complex research workflows. Alongside these, academic and open-source initiatives have made significant structural leaps. Frameworks like STORM \cite{shao2024assisting} simulate multi-perspective discovery and drafting processes to generate Wikipedia-like articles, while systems like Open Deep Research leverage underlying search mechanisms to autonomously orchestrate extensive literature reviews and cross-reference multiple sources. To manage these immense cognitive loads, modern research architectures often incorporate multi-agent debate protocols (e.g., AutoGen \cite{wu2024autogen}) and dynamic context management to maintain coherence. Yet, as both Deep Search and Deep Research frameworks transition toward these highly autonomous, long-horizon workflows, they become increasingly susceptible to compounding cognitive bottlenecks. Despite these architectural leaps, the field still lacks a systematic, granular understanding of how and why these agents fail during their intermediate exploration and synthesis phases.

\subsection{Evaluation Paradigms of Deep Search and Deep Research Agents}

To evaluate the effectiveness of Web autonomous systems, the research community has developed a diverse set of benchmarks targeting web navigation and open-ended knowledge synthesis.  For Deep Search capabilities, early multi-hop question answering datasets such as HotpotQA \cite{yang2018hotpotqa} established foundational settings for evaluating complex retrieval and reasoning. Subsequent benchmarks extend this paradigm into interactive, web-grounded environments, including Mind2Web \cite{deng2023mind2web}, WebArena \cite{zhou2023webarena}, and BrowseComp \cite{wei2025browsecomp}, where agents must execute long-horizon navigation over real web environment. In parallel, evaluation of Deep Research frameworks has shifted toward large-context integration and long-form synthesis. Benchmarks such as Deep Research Bench and ADR-Bench assess an agent's ability to aggregate information across extensive sources and produce coherent analytical reports.

While these benchmarks have driven significant architectural progress, they are fundamentally constrained by their \textbf{outcome-centric evaluation paradigms}. Existing methodologies predominantly assess agents based on end-to-end metrics: binary task success rates, exact string matches for factual queries, or LLM-as-a-Judge holistic scoring of the final synthesized report. This reliance on terminal outcomes creates a substantial evaluation blind spot. A correct final answer often masks severe process-level flaws, while an incorrect outcome provides no actionable insight into the specific cognitive or navigational bottleneck. Particularly in Deep Search and Deep Research settings, where agent trajectories span dozens of steps and consume large context windows, intermediate failures are poorly reflected by aggregate outcome metrics, making error attribution difficult despite low final accuracy scores. Our work directly challenges this opaque paradigm. By shifting the evaluation focus from terminal outcome metrics to step-level diagnostic analysis, we enable finer-grained analysis of failures that are difficult to observe through aggregate benchmark scores alone.

\subsection{Error Analysis and Diagnostics in LLMs}

Prior work on LLM error analysis has primarily examined failures arising in static generation and localized reasoning settings. Early diagnostic studies investigated factual hallucinations and model reliability, analyzing how language models produce unsupported or fabricated content under knowledge-intensive queries \cite{ji2023survey,kadavath2022language}. Complementary lines of work further examined reasoning inconsistencies in Chain-of-Thought (CoT) outputs, revealing mismatches between intermediate reasoning traces and final answers, as well as limitations in compositional generalization \cite{gendron2023large,press2023measuring}.

As language models have evolved into interactive agents, analytical attention has increasingly shifted toward failures arising during tool use and environment interaction. Recent studies conduct systematic error analyses of tool-augmented LLMs, identifying recurrent execution issues such as invalid action formatting, incorrect argument grounding, and inconsistent tool-selection behaviors \cite{zhu2025llm}. While these analyses provide valuable insights into general-purpose agent behavior across broad task settings, they place less emphasis on information-seeking systems centered on iterative retrieval and large-scale evidence synthesis. Failure modes specific to Deep Search and Deep Research workflows therefore remain comparatively underexplored. Our work introduces a taxonomy aimed at more systematically characterizing failure patterns in long-horizon, information-dense autonomous research workflows.

\section{Dataset}
\label{app:dataset_details}

This appendix summarizes the dataset scale used in our evaluation.

For Deep Search, we use four benchmarks: BrowseComp, BrowseComp-zh, GAIA-val-text, and XBench-DS. Due to evaluation cost constraints, BrowseComp and BrowseComp-zh are evaluated using subsampled subsets.

For Deep Research, we use DeepResearch Bench as the evaluation source for long-horizon synthesis tasks.

\begin{table*}[!htbp]
\centering
\caption{Dataset statistics used in this work.}
\label{tab:dataset_stats_appendix}
\begin{tabular*}{0.95\textwidth}{@{}l l c c l@{}}
\toprule
\textbf{Benchmark} & \textbf{Paradigm} & \textbf{Language} & \textbf{\#Tasks} & \textbf{Primary Use} \\
\midrule
BrowseComp & Deep Search & English & 100 & Multi-hop web retrieval/navigation \\
BrowseComp-zh & Deep Search & Chinese & 100 & Multilingual web retrieval/navigation \\
GAIA (val-text-only) & Deep Search & English & 103 & Tool-assisted textual reasoning \\
XBench-DS & Deep Search & Chinese & 100 & Iterative query reformulation/retrieval \\
DeepResearch Bench & Deep Research & English, Chinese & 100 & Long-horizon synthesis/report generation \\
\bottomrule
\end{tabular*}%

\end{table*}

\textbf{Success Criterion.} For Deep Search, success is defined by answer correctness against the ground truth. Given model output $\hat{y}$ and reference answer $y^*$, a run is considered successful when $\hat{y}$ matches $y^*$ under benchmark normalization rules (e.g., alias normalization and formatting tolerance). This criterion is relatively uniform across Deep Search benchmarks because these tasks typically have explicit reference answers.

 For Deep Research, success criteria are benchmark-dependent rather than uniform. In this work, we follow the DeepResearch Bench \cite{du2025deepresearch} setting and evaluate along two dimensions:
\begin{itemize}
    \item \textbf{RACE} (Reference-based and Adaptive Criteria-driven Evaluation)
    \begin{itemize}
        \item \textbf{Comprehensiveness}: coverage breadth and depth of the research topic.
        \item \textbf{Insight/Depth}: quality of analysis and insight generation.
        \item \textbf{Instruction-Following}: adherence to specific task requirements.
        \item \textbf{Readability}: clarity, organization, and presentation quality.
    \end{itemize}
    \item \textbf{FACT} (Factual Abundance and Citation Trustworthiness)
    \begin{itemize}
        \item \textbf{Citation Accuracy}: percentage of correctly supported citations.
        \item \textbf{Effective Citations}: average number of verifiably supported citations per task.
    \end{itemize}
\end{itemize}

\section{Prompt}
\label{app:prompts}

\begin{promptbox}[System Prompt for System-Level Error Diagnosis]
\begin{Verbatim}[breaklines=true, breakanywhere=true, breaksymbolleft=, breaksymbolright=, fontsize=\small]
"""
You are an expert AI Agent Behavior Auditor. Your goal is to analyze the internal reasoning and tool-calling behavior of an AI Agent to identify specific error patterns based on provided logs and standard specifications.

## Input Data
1. **Think Fragment**: {{THINK_CONTENT}}
2. **Tool Call Details**: {{TOOL_DETAILS}}

## Standard Tool Specifications

{'name': 'tool-reasoning', 
    'tools': [{'name': 'reasoning', 'description': 'This tool is for pure text-based reasoning, analysis, and logical thinking. It integrates collected information, organizes final logic, and provides planning insights.\n\nIMPORTANT: This tool cannot access the internet, read files, program, or process multimodal content. It only performs pure text reasoning.\n\nUse this tool for:\n- Integrating and synthesizing collected information\n- Analyzing patterns and relationships in data\n- Logical reasoning and problem-solving\n- Planning and strategy development\n- Complex math problems, puzzles, riddles, and IQ tests\n\nDO NOT use this tool for simple and obvious questions.\n\nArgs:\n    question: The complex question or problem requiring step-by-step reasoning. Should include all relevant information needed to solve the problem.\n\nReturns:\n    The reasoned answer to the question.', 'schema': {'properties': {'question': {'title': 'Question', 'type': 'string'}}, 'required': ['question'], 'type': 'object'}}]}, 
{'name': 'tool-searching', 
    'tools': [
    {'name': 'google_search', 'description': \"Perform google searches via Serper API and retrieve rich results.\\nIt is able to retrieve organic search results, people also ask, related searches, and knowledge graph.\\n\\nArgs:\\n    q: Search query string.\\n    gl: Country context for search (e.g., 'us' for United States, 'cn' for China, 'uk' for United Kingdom). Influences regional results priority. Default is 'us'.\\n    hl: Google interface language (e.g., 'en' for English, 'zh' for Chinese, 'es' for Spanish). Affects snippet language preference. Default is 'en'.\\n    location: City-level location for search results (e.g., 'SoHo, New York, United States', 'California, United States').\\n    num: The number of results to return (default: 10).\\n    tbs: Time-based search filter ('qdr:h' for past hour, 'qdr:d' for past day, 'qdr:w' for past week, 'qdr:m' for past month, 'qdr:y' for past year).\\n    page: The page number of results to return (default: 1).\\n\\nReturns:\\n    The search results.\", 'schema': {'properties': {'q': {'title': 'Q', 'type': 'string'}, 'gl': {'default': 'us', 'title': 'Gl', 'type': 'string'}, 'hl': {'default': 'en', 'title': 'Hl', 'type': 'string'}, 'location': {'default': None, 'title': 'Location', 'type': 'string'}, 'num': {'default': 10, 'title': 'Num', 'type': 'integer'}, 'tbs': {'default': None, 'title': 'Tbs', 'type': 'string'}, 'page': {'default': 1, 'title': 'Page', 'type': 'integer'}}, 'required': ['q'], 'type': 'object'}}
    {'name': 'wiki_get_page_content', 'description': 'Get specific Wikipedia page content for the specific entity (people, places, concepts, events) and return structured information.\\n\\nThis tool searches Wikipedia for the given entity and returns either the first few sentences\\n(which typically contain the summary/introduction) or full page content based on parameters.\\nIt handles disambiguation pages and provides clean, structured output.\\n\\nArgs:\\n    entity: The entity to search for in Wikipedia.\\n    first_sentences: Number of first sentences to return from the page. Set to 0 to return full content. Defaults to 10.\\n\\nReturns:\\n    str: Formatted search results containing title, first sentences/full content, and URL.\\n         Returns error message if page not found or other issues occur.', 'schema': {'properties': {'entity': {'title': 'Entity', 'type': 'string'}, 'first_sentences': {'default': 10, 'title': 'First Sentences', 'type': 'integer'}}, 'required': ['entity'], 'type': 'object'}}
    {'name': 'search_wiki_revision', 'description': 'Search for an entity in Wikipedia and return the revision history for a specific month.\\n\\nArgs:\\n    entity: The entity to search for in Wikipedia.\\n    year: The year of the revision (e.g. 2024).\\n    month: The month of the revision (1-12).\\n    max_revisions: Maximum number of revisions to return. Defaults to 50.\\n\\nReturns:\\n    str: Formatted revision history with timestamps, revision IDs, and URLs.\\n         Returns error message if page not found or other issues occur.', 'schema': {'properties': {'entity': {'title': 'Entity', 'type': 'string'}, 'year': {'title': 'Year', 'type': 'integer'}, 'month': {'title': 'Month', 'type': 'integer'}, 'max_revisions': {'default': 50, 'title': 'Max Revisions', 'type': 'integer'}}, 'required': ['entity', 'year', 'month'], 'type': 'object'}}
    {'name': 'search_archived_webpage', 'description': 'Search the Wayback Machine (archive.org) for archived versions of a webpage, optionally for a specific date.\\n\\nArgs:\\n    url: The URL to search for in the Wayback Machine.\\n    year: The target year (e.g., 2023).\\n    month: The target month (1-12).\\n    day: The target day (1-31).\\n\\nReturns:\\n    str: Formatted archive information including archived URL, timestamp, and status.\\n         Returns error message if URL not found or other issues occur.', 'schema': {'properties': {'url': {'title': 'Url', 'type': 'string'}, 'year': {'title': 'Year', 'type': 'integer'}, 'month': {'title': 'Month', 'type': 'integer'}, 'day': {'title': 'Day', 'type': 'integer'}}, 'required': ['url', 'year', 'month', 'day'], 'type': 'object'}}, 
    {'name': 'scrape_website', 'description': 'This tool is used to scrape a website for its content. Search engines are not supported by this tool. This tool can also be used to get YouTube video non-visual information (however, it may be incomplete), such as video subtitles, titles, descriptions, key moments, etc.\\n\\nArgs:\\n    url: The URL of the website to scrape.\\nReturns:\\n    The scraped website content.', 'schema': {'properties': {'url': {'title': 'Url', 'type': 'string'}}, 'required': ['url'], 'type': 'object'}}]}
{'name': 'tool-reading', 
    'tools': [
        {'name': 'read_file', 'description': \"Read various types of resources (Doc, PPT, PDF, Excel, CSV, ZIP file etc.)\\ndescribed by an file: or data: URI.\\n\\nArgs:\\n    uri: Required. The URI of the resource to read. Need to start with 'file:' or 'data:' schemes. Files from sandbox are not supported. You should use the local file path.\\n\\nReturns:\\n    str: The content of the resource, or an error message if reading fails.\", 'schema': {'properties': {'uri': {'title': 'Uri', 'type': 'string'}}, 'required': ['uri'], 'type': 'object'}}]}]

---

## Error Taxonomy
Please evaluate the interaction against the following categories:

### 1. Agent Logic & Decision Errors
- **Tool Call Parsing Failure**: The Agent failed to output valid syntax for the tool environment. **Crucial Example**: Unclosed tags (e.g., `<mcp-server>` without a closing `</mcp-server>`) or malformed structural markers. This is an Agent error, NOT a tool argument error.
- **Incorrect Tool Name**: The Agent attempts to call a tool that does not exist in the specifications (e.g., typos or hallucinations). 

### 2. Search Tool Errors
- **Call Formatting Error**: Missing required arguments or incorrect data types.
- **Tool-Function Mismatch**: Using a search query field for tasks like URL scraping (e.g., q="http://...").
- **Irrelevant Query**: The query is meta-talk or non-functional (e.g., q="scrape website" or q="find info").
- **Domain Tool Misalignment**: Using a general tool when a specific one is required (e.g., searching Wikipedia for a known Arxiv ID).
- **Syntatical Query Failure**: Query contains excessive special characters or "hallucinated" syntax that results in zero hits.

### 3. Read/Scrape Errors
- **Unsupported Content Type**: Attempting to "read" multimodal data (images/video) or unsupported file formats.
- **Non-Existent Page**: The target URL/Path is invalid or returns a 404 error.
- **Anti-Scraping Blockers**: The result indicates the Agent was blocked by a WAF, CAPTCHA, or "Access Denied" page.

### 4. Discovery: New Error Patterns
- **Undefined Error**: If you identify a behavior that is clearly erroneous, suboptimal, or logically flawed but **does not fit into any of the categories above**, you MUST report it as a "New Error Pattern", and try to explain what type it is.

---

## Output Instructions
Analyze the inputs step-by-step. 
1. **Validation**: Compare the "Tool Call Details" against the "Standard Tool Specifications."
2. **Analysis**: If the reasoning, tool choice, and parameters are correct and logical, and the execution result (if provided) shows no failure, classify the interaction as correct.
3. **Error Reporting**: If an error is found, identify the category. If it is an error not listed above, report it as a "New Error Pattern".

**Output Format (JSON):**
```json
{
  "is_error": true/false,
  "status": "Correct / Error Detected",
  "pattern_category": "Category - Sub-item (or 'No Error' or 'New Error Pattern')",
  "is_new_pattern": true/false, 
  "evidence": "Snippet from the input supporting the status",
  "analysis": "Explanation of why the interaction is correct or why it is considered an error."
}
"""

\end{Verbatim}
\end{promptbox}

\begin{promptbox}[System Prompt for Task-Level Error Diagnosis]
\begin{Verbatim}[breaklines=true, breakanywhere=true, breaksymbolleft=, breaksymbolright=, fontsize=\small]


"""
You are an expert evaluator auditing the intermediate execution traces of a Deep Search Agent. 
Your objective is to evaluate whether the agent successfully retrieved the complete information required to answer the query.

Analyze the provided [USER QUERY], the [GROUND TRUTH] answer, and the [CLEAN EXECUTION TRANSCRIPT].

# TAXONOMY OF DEEP SEARCH ERRORS:
1. SUCCESS: The agent formulated effective queries, explored adequately, and successfully retrieved all necessary context matching the Ground Truth.
2. INFORMATION_MISSING: The agent completely failed to retrieve the core information, or the transcript lacks crucial details present in the Ground Truth.
3. PREMATURE_STOPPING: The agent stopped searching before gathering all elements of the Ground Truth (e.g., in exhaustive counting tasks).
4. RABBIT_HOLE_TRAP: The agent wasted multiple steps (3+ steps) clicking through irrelevant sub-domains, or got stuck in an infinite loop.
5. INEFFECTIVE_QUERY: The agent used poorly formulated search queries or repeatedly submitted the exact same query despite no useful results.

# INSTRUCTIONS:
- Compare the facts retrieved in the [CLEAN EXECUTION TRANSCRIPT] against the facts explicitly stated in the [GROUND TRUTH].
- Track the agent's logic step-by-step.
- Provide reasoning diagnosing the search process before outputting the final classification.
"""

\end{Verbatim}
\end{promptbox}

\section{Case Studies}
To complement aggregate statistics, we present representative case studies from Deep Search and Deep Research. Each case highlights how intermediate failures emerge, propagate, and affect final outcomes, illustrating the practical value of our process-level taxonomy.

\subsection{System-Level Error}

\begin{tcolorbox}[colback=gray!5!white,breakable,colframe=gray!50!black,title=Tool-Function Mismatch]
\small
\textbf{Thought:} \textit{``Open the page.''} \\
\textbf{Tool Call:} \texttt{google\_search} \\
\textbf{Arguments:}\texttt{\{"q": "https://www.history.nav\\y.mil/.../allied-ships.html"\}}\\
\textbf{Diagnosis:} The agent's intent was to open and read a specific webpage. However, it invoked the \texttt{google\_search} tool using the full URL as the query. This explicitly mismatches the tool's intended purpose, as \texttt{google\_search} is designed to return search result snippets rather than full DOM content. The correct tool for this operation is \texttt{scrape\_website} via the \texttt{url} argument. Although the invocation format is syntactically flawless, the improper tool selection directly constitutes a \textit{Tool-Function Mismatch}, demonstrating a gap in the agent's environmental grounding. \\
\textbf{Result:} Tool-Function Mismatch
\end{tcolorbox}

\vspace{1em}

\begin{tcolorbox}[colback=gray!5!white,breakable,colframe=gray!50!black,title=Call Formatting Error]
\small
\textbf{Thought:} \textit{``Now open the page.''} \\
\textbf{Tool Call:} \texttt{google\_search} \\
\textbf{Arguments:} \texttt{\{"q": "Bull Wall Wikipedia page content", "gl": "ie", "hl": "en", "num": 100, "first\_sentences": 0\}} \\
\textbf{Diagnosis:} The agent's stated intent was to open and read a specific webpage. However, it mistakenly invoked the \texttt{google\_search} tool instead of a targeted scraping tool. More critically, the agent hallucinated the argument \texttt{"first\_sentences": 0}. This parameter is entirely absent from the \texttt{google\_search} schema but is typically associated with Wikipedia-specific retrieval tools (e.g., \texttt{wiki\_get\_page\_content}). This cross-contamination of tool schemas results in an immediate API rejection. The failure highlights a severe environmental confusion where the agent conflates the parameters of distinct tools, ultimately triggering a strict \textit{Call Formatting Error} error. \\
\textbf{Result:} Call Formatting Error
\end{tcolorbox}
\vspace{1em}

\subsection{Task-Level Error}

\begin{tcolorbox}[colback=blue!5!white,breakable,colframe=blue!40!black,title=Deep Search: Ineffective Query]
\small

\textbf{Trajectory Summary:} The agent never retrieved concrete evidence from web pages and failed to identify any candidate band or video. It cycled through broad, unfocused, and increasingly speculative queries (e.g., generic searches for bands with 'Saint' in the name, music videos about mother–child grief, various national tragedies) without drilling into specific sources or scraping actual pages. No substantive observations were collected that could be compared to the Ground Truth ('The Light'). There is no sign of ignoring retrieved evidence (so not OBSERVATION\_NEGLECT), nor reliance on snippets (SURFACE\_LEVEL\_SKIMMING) because no snippets or scraped content were presented. The session ended with the agent acknowledging failure, having relied on poorly targeted queries and conjecture rather than effective information retrieval. Hence, the primary failure is INEFFECTIVE\_QUERY at the initial search stage. \\
\textbf{Error Category:} \texttt{INEFFECTIVE\_QUERY} \\
\end{tcolorbox}

\begin{tcolorbox}[colback=teal!5!white,breakable,colframe=teal!50!black,title=Deep Research: Synthesis Hallucination{,} Contradiction Neglect and Sourcing Bias]
\small

\textbf{Prompt:} Find data and evidence to support or refute the hypothesis that an airport handling an annual passenger throughput of 500,000 (five hundred thousand) can generate significant and measurable socioeconomic impacts on its surrounding region. \\

\text{Contradiction Neglect and Sourcing Bias}: The agent presented conflicting quantitative figures without reconciliation. For Rapid City Regional Airport, a search snippet indicated direct spending of ~\$295M in 2023 (Table 19), yet the report cites \$153M direct and \$188M total impact for ~338k passengers without cross-validating or explaining the discrepancy. The job-per-seat calculation is also internally inconsistent: after quoting 0.0114 jobs per departing seat (non-hub) and 0.005 (small hub), the agent derives 182.5 jobs from 36,500 seats (which matches 0.005, not 0.0114) and later scales 500k seats to ~1,560 jobs (which doesn’t match either multiplier). These contradictions were neither flagged nor reconciled.     \\
\textbf{Synthesis Hallucination}: Many cited sources were never successfully accessed (e.g., PSP economic impact report, RAP report PDF, Dane County study, EU SGEI CMS page, ACI Europe connectivity elasticities), yet the agent provided precise figures (e.g., PSP: 18,377 jobs, \$1.5B GDP, \$2.5B output; Dane County: 5,512 direct jobs, \$425.3M wages; ACI Europe: 10\% connectivity -> 0.5\% GDP and 1.6\% employment) and detailed scaling estimates. The agent also used internal placeholders (web\_xxxxx) as if they were verified citations. This indicates the agent filled a complex structure with unverified or invented data rather than acknowledging access failures or uncertainty. \\
\textbf{Goal Drifting}: Despite sourcing issues, the final report remained focused on the original task: evaluating whether a 500,000-passenger airport can generate significant, measurable socioeconomic impacts. It presented supportive evidence, counter-evidence, methodological notes, and a synthesized conclusion addressing the hypothesis. \\
\textbf{Error Categories:} \texttt{Synthesis Hallucination}, \texttt{Contradiction Neglect and Sourcing Bias}\\
\end{tcolorbox}

\end{document}